\documentclass[sigconf, screen]{acmart}
\usepackage{amsmath,amsfonts}
\usepackage{algorithmic}
\usepackage{algorithm}
\usepackage{array}
\usepackage[caption=false,font=normalsize,labelfont=sf,textfont=sf]{subfig}
\usepackage{stfloats}
\usepackage{url}
\usepackage{graphicx}
\usepackage{booktabs}
\usepackage{pifont}
\usepackage{multirow}
\usepackage{enumitem}
\usepackage{stfloats}
\usepackage{caption}
\usepackage{makecell}
\usepackage{lineno}
\usepackage{xspace}
\usepackage{boldline}

\definecolor{citecolor}{RGB}{0, 0, 255}  
\definecolor{mygray}{gray}{0.9}

\makeatletter

\newcommand{\thickhline}{%
	\noalign {\ifnum 0=`}\fi \hrule height 1pt
	\futurelet \reserved@a \@xhline
}

\DeclareRobustCommand\onedot{\futurelet\@let@token\@onedot}
\def\@onedot{\ifx\@let@token.\else.\null\fi\xspace}

\def\eg{\emph{e.g}\onedot} 
\def\ie{\emph{i.e}\onedot}

\def\BibTeX{{\rm B\kern-.05em{\sc i\kern-.025em b}\kern-.08em
		T\kern-.1667em\lower.7ex\hbox{E}\kern-.125emX}}

\AtBeginDocument{%
  \providecommand\BibTeX{{%
    Bib\TeX}}}

\copyrightyear{2025}
\acmYear{2025}
\setcopyright{acmlicensed}
\acmConference[MM '25] {Proceedings of the 33rd ACM International Conference on Multimedia}{October 27--31, 2025}{Dublin, Ireland.}
\acmBooktitle{Proceedings of the 33rd ACM International Conference on Multimedia (MM '25), October 27--31, 2025, Dublin, Ireland}
\acmISBN{979-8-4007-2035-2/2025/10}
\acmDOI{10.1145/XXXXXX.XXXXXX}

\begin{document}

\title{\textsc{Lava}: Language Driven Scalable and Versatile Traffic Video Analytics}

\author{Yanrui Yu}
\affiliation{%
	\institution{Beijing Institute of Technology}
	\city{Beijing}
	\country{China}
}
\email{yanruiyu@bit.edu.cn}

\author{Tianfei Zhou$^\dagger$}
\affiliation{
	\institution{Beijing Institute of Technology}
	\city{Beijing}
	\country{China}
}
\email{tfzhou@bit.edu.cn}

\author{Jiaxin Sun}
\affiliation{
	\institution{Beijing Institute of Technology}
	\city{Beijing}
	\country{China}
}
\email{sunjiaxin@bit.edu.cn}

\author{Lianpeng Qiao}
\affiliation{
	\institution{Beijing Institute of Technology}
	\city{Beijing}
	\country{China}
}
\email{qiaolp@bit.edu.cn}

\author{Lizhong Ding}
\affiliation{
	\institution{Beijing Institute of Technology}
	\city{Beijing}
	\country{China}
}
\email{lizhong.ding@outlook.com}

\author{Ye Yuan$^\dagger$}
\affiliation{
	\institution{Beijing Institute of Technology}
	\city{Beijing}
	\country{China}
}
\email{yuan-ye@bit.edu.cn}

\author{Guoren Wang}
\affiliation{
	\institution{Beijing Institute of Technology}
	\city{Beijing}
	\country{China}
}
\email{wanggr@bit.edu.cn}

\renewcommand{\shortauthors}{Yanrui Yu, Tianfei Zhou, \& Jiaxin Sun }

\begin{abstract}
In modern urban environments, camera networks generate massive amounts of operational footage -- reaching petabytes each day -- making scalable video analytics essential for efficient processing.  Many existing approaches adopt an SQL-based paradigm for querying such large-scale video databases; however, this constrains queries to rigid patterns with predefined semantic categories,  significantly limiting analytical flexibility.  In this work, we explore a  language-driven video analytics paradigm aimed at enabling flexible and efficient querying of high-volume video data driven by natural language. Particularly, we build \textsc{Lava}, a system that accepts natural language queries and  retrieves traffic targets across multiple levels of granularity and arbitrary categories. \textsc{Lava} comprises three main components: 1) a multi-armed bandit-based efficient sampling method for video segment-level localization;
 2) a video-specific open-world detection module for object-level retrieval; and 3) a long-term object trajectory extraction scheme for temporal object association, yielding complete trajectories for object-of-interests. To support comprehensive evaluation, we further develop a novel benchmark by providing diverse, semantically rich natural language predicates and fine-grained annotations for multiple videos. Experiments on this benchmark demonstrate that \textsc{Lava} improves $F_1$-scores for selection queries by $\mathbf{14\%}$, reduces MPAE for aggregation queries by $\mathbf{0.39}$, and achieves top-$k$ precision of $\mathbf{86\%}$, while processing videos $ \mathbf{9.6\times} $ faster than the most accurate baseline. Our code and dataset are available at \url{https://github.com/yuyanrui/LAVA}.
\end{abstract}



\maketitle

\vspace{-5pt}
\let \mybackup \thefootnote
\let \thefootnote \relax
\footnotetext{$^\dagger$Corresponding author.}
\let \thefootnote \mybackup
\let \mybackup \imareallyundefinedcommand

\begin{figure}
	\centering
	\includegraphics[width=\linewidth]{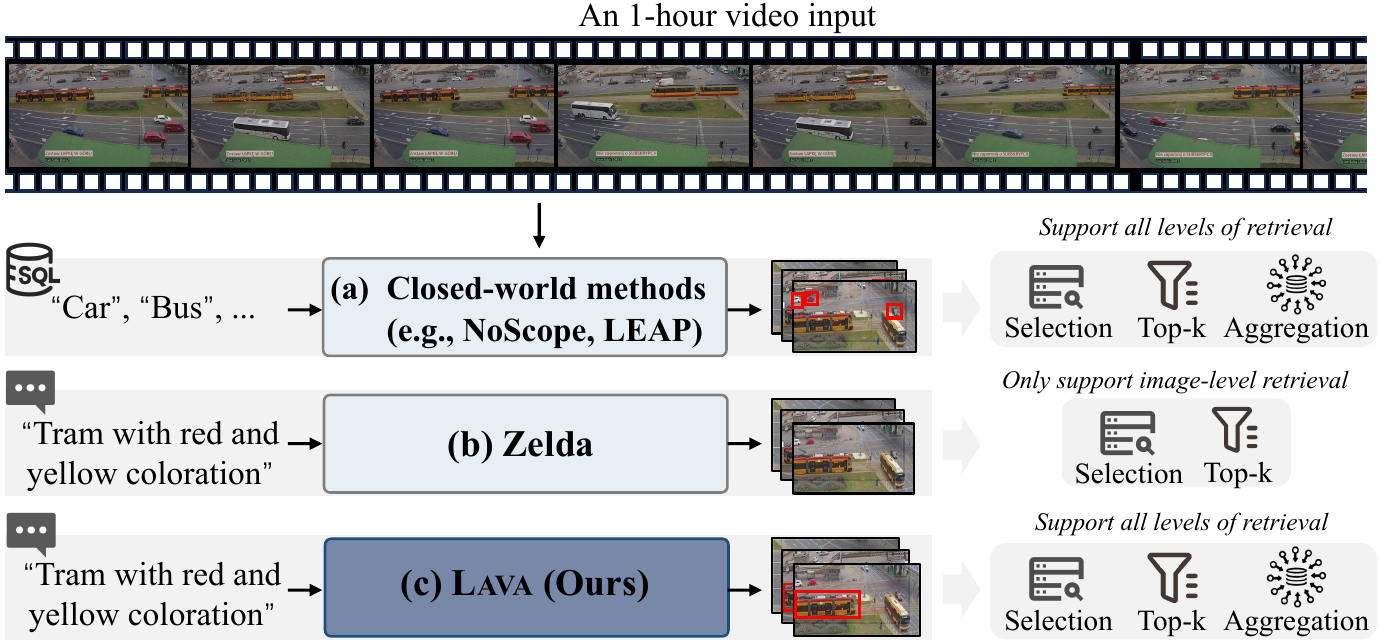}
	\caption{\small{(a) Traditional scalable video analytics methods (\eg, NoScope \cite{kang2017noscope}, LEAP \cite{xu2024predictive}, BlazeIt \cite{kang13blazeit}, OTIF \cite{bastani2022otif}) are SQL-based, closed-world systems, limited to querying predefined object classes (\eg, car, bus). (b) Zelda \cite{romero2023zelda} is a recent effort towards open-world video analytics using  language queries but is constrained to image-level query. (c)  Our system, \textsc{Lava}, tackles all these limitations, and supports versatile video analytics tasks -- including selection,  top-$k$, aggregation -- using natural language query predicates. }}
	\label{fig: motivation}
	\vspace{-15pt}
\end{figure}

\section{Introduction}
\label{sec:introduction}

The exponential growth of traffic video data presents significant analytical challenges. For example,  Shanghai's 40,000 surveillance cameras produces approximately 1.68PB videos each day \cite{multiagent}. While these data are crucial for  applications such as real-time traffic monitoring, accident detection, and infrastructure optimization \cite{dong2019feature, koudas2020video}, their scale overwhelms infrastructure and dynamic complexity surpasses systems restricted to predefined taxonomies \cite{liang2020video, cao2023video}. To handle such large-scale video database,  scalable video analytics systems \cite{kang2017noscope,kang13blazeit,bastani2022otif,romero2023zelda} are developed, which typically support three fundamental query types: 1) \textbf{\textit{selection}} to pinpoint frames matching specific predicates (\eg, identifying frames including cars), 2) \textit{\textbf{top-$k$}} to retrieve the $k$ best-matching frames (\eg the top‑$5$ frames containing the most number of cars) and 3) \textit{\textbf{aggregation}}  to compute statistical summaries over the entire video (\eg calculating the average number of cars per frame).


Recent advancements in scalable video analytics evolve through two complementary strategies: proxy models reduce computational demands by replacing costly DNNs with lightweight alternatives \cite{kang2017noscope,kang13blazeit,lai2021top,cao2022figo,bastani2022otif} to filter irrelevant frames, while adaptive sampling methods \cite{bastani2020miris,xu2024predictive} exploit temporal patterns to minimize redundancy—both operating under closed-world paradigms limited to predefined object categories, as illustrated in Fig. \ref{fig: motivation} (a). To transcend these limitations, vision-language models (VLMs) like \cite{romero2023zelda,moll2023seesaw} enable open-world semantic queries via natural language (\eg, ``cars at intersections''), yet remain confined to image-level analysis without object-centric temporal operations (\eg, counting ``trams during peak hours''), as shown in Fig. \ref{fig: motivation} (b). Meanwhile, recent large video models  \cite{li2023videochat,wang2024qwen2,team2024gemini,zhang2024video} demonstrate impressive grounding and multimodal understanding aligned with natural language prompts. However, their prohibitive resource requirements render them impractical for high-volume, hour-long videos -- for instance, Qwen2-VL requires 40GB GPU memory and 90 hours to process a single one-hour video on an A100 GPU. 
In contrast, our work focuses on efficient, scalable analysis of high-volume video data, bridging the gap between open-world expressivity and practical deployment.

We propose a new paradigm, \textbf{language-driven scalable video analytics}, which enables users to retrieve semantically rich video content using natural language, so as to meet open-world query requirements. This paradigm has two key features as shown in Fig. \ref{fig: motivation} (c): \textit{1) Query semantically rich content}: Users can describe intricate details, attributes, and relationships of objects or scenes in natural language, enabling expressive and flexible queries.  \textit{2) Query with diverse methods}: Supports selection, aggregation, and top-$k$ queries for multi-granular analysis (\eg, frame-level selection, object-specific aggregation) and diverse analytical tasks. Despite impressive, new challenges arise due to the inherent nature of natural language query predicates and the complexities they introduce in video analytics:

\noindent {\ding{182}\hspace{0.5em}\textbf{Target Sparsity:}} In high-volume, hour-long video scenarios, query targets typically appear in only a small fraction of frames—\eg, the ``Black pickup truck'' appears for just 2.4 minutes in a one-hour video in the Caldot2 \cite{bastani2022otif}. This extreme sparsity renders exhaustive frame-by-frame analysis highly inefficient, imposing substantial computational costs when analyzing high-volume video data.

\noindent {\ding{183}\hspace{0.5em} \textbf{Open-World:}} Language-driven video analytics struggle with queries beyond predefined categories. Open-vocabulary object detectors, built on visual-language pre-training, seem to be a feasible way to manage this. However, existing solutions  \cite{cheng2024yolo, 23Grounding} often suffer from  low precision and recall in open-world settings, confusing similar objects (\eg, ``red car'' vs. ``blue car'') and having difficulty under complicated conditions like occlusion or lighting changes. 

\noindent {\ding{184}\hspace{0.5em} \textbf{Lack of Benchmark Datasets:}} Existing datasets, such as those from BlazeIt \cite{kang13blazeit}, MIRIS \cite{bastani2020miris} and OTIF \cite{bastani2022otif}, are annotated with predefined classes like “car”, “bus”, and “truck”, limiting their ability to capture the diversity and complexity of natural language predicates. This lack of semantic richness impedes the development and evaluation of systems for language queries in video analytics.


To address these challenges, we present \textsc{\textbf{Lava}}, a language-driven scalable video analytics system  for traffic scenarios, which produce petabytes of data daily and involve open-world complexities such as various vehicle types and transient interactions requiring detailed semantic interpretation. Particularly,
to address \ding{182}, \textsc{Lava} introduces a multi-armed bandit-based relevant segment localization module to efficiently localize video segments that are likely to contain query targets. It models each equal-length segment as an ``arm'' in a multi-armed bandit framework, and applies a probabilistic exploration-exploitation strategy to focus processing on promising segments, significantly reducing computational demands.
To overcome \ding{183},  we propose video-specific open-world object detection and long-term object trajectory extraction. The former leverages CLIP, which is tuned specifically for each video, to extract objects aligning with the query. Subsequently, leveraging the fixed-viewpoint nature of traffic scenes, the latter assigns motion patterns to associate detections temporally, efficiently mapping complete trajectories of query-relevant objects, thereby enhancing trajectory completeness over long spans. To address \ding{184}, we develop a benchmark tailored for language-driven video analytics. Building upon six datasets proposed in prior works \cite{bastani2020miris,kang13blazeit,bastani2022otif}, we introduce 18 semantically rich, natural language query predicates, allowing comprehensive evaluation of systems such as \textsc{Lava}.

Leveraging its innovative strategies, \textsc{Lava} demonstrates outstanding performance in both efficiency and accuracy on our newly developed benchmark. Specifically, it delivers a \textbf{15\% improvement} in $F_1$-score for selection queries, \textbf{reduces MPAE by 0.39} for aggregation queries, and operates \textbf{9.6$\times$ faster} than the most accurate baseline. These results, comprehensively evaluated on our benchmark, highlight \textsc{Lava}'s ability to deliver \textbf{state-of-the-art performance in query accuracy and computational efficiency}, setting a new standard for language-driven video analytics.
\vspace{-5pt}

\section{Related Work}
\subsection{Scalable Video Analytics}
Traditional scalable video analytics methods primarily focus on improving query efficiency by leveraging sampling strategies and proxy models to balance accuracy and speed \cite{kang2017noscope, kang13blazeit, hsieh2018focus, cao2022figo, bastani2020miris, bastani2022otif, kang2022viva, xu2024predictive,yuan2024nsdb}. NoScope \cite{kang2017noscope} utilizes cascades of lightweight models tailored for specific queries in fixed-angle video streams, achieving up to 1000× speed-ups by filtering redundant frames. BlazeIt \cite{kang13blazeit} optimizes queries by leveraging approximate representations for tasks like aggregation. FiGO \cite{cao2022figo} introduces fine-grained optimization to dynamically adjust model selection based on video complexity. OTIF \cite{bastani2022otif} integrates multi-object tracking (MOT) and frame down-sampling to accelerate query responses. LEAP \cite{xu2024predictive} reduces frame processing using predictive sampling and motion pattern management, making it effective for high-throughput video analytics. While these methods achieve significant computational savings, their reliance on predefined classes limits their flexibility, making them unsuitable for expressive, language-driven queries.

The integration of vision-language models (VLMs) into video analytics has enabled zero-shot natural language queries that bypass predefined categories. Zelda \cite{romero2023zelda} employs VLMs to process natural language queries, enhancing flexibility by eliminating the need for task-specific training. Zelda also addresses redundancy in VLM outputs to improve query diversity. However, Zelda is restricted to image-level queries, limiting its applicability to detailed video tasks, such as temporal aggregation. Furthermore, its reliance on out-of-the-box VLMs reduces its ability to handle nuanced queries, often resulting in semantically inconsistent outputs.

Our method overcomes existing system limitations by supporting flexible, natural language-driven queries that exceed predefined categories and limited query types. Unlike traditional fixed-class constraints or image-level selection systems, our approach handles diverse queries (selection, aggregation, top-$k$) . This enables detailed, precise analysis, especially suited for complex real-world scenarios.

\subsection{Large Video Models}
Large Video Models have made significant progress in video grounding by bridging vision encoders \cite{dosovitskiy2020image, radford2021learning, zhai2023sigmoid} and language models via spatiotemporal adapters \cite{li2023blip, liu2023visual}. Methods such as VideoChat \cite{li2023videochat}, LLaVA-Video \cite{zhang2024video}, GPT-4o \cite{hurst2024gpt}, Gemini 1.5 Pro \cite{team2024gemini}, Video-Llama 2 \cite{damonlpsg2024videollama2}, Qwen2-VL \cite{wang2024qwen2} and others \cite{song2024moviechat, zhang2024long, liu2024kangaroo, luo2024video, wang2024videotree, tang2025adaptive, wang2025videorft} leverage large-scale training to optimize multimodal alignment, achieving impressive performance on video grounding. However, despite their success, these methods face inherent limitations when processing high-volume video data. In video analytics, it is essential to extract detailed frame-level information to determine whether each frame contains objects relevant to the query. Although processing videos frame-by-frame using MLLMs is theoretically feasible, such a strategy becomes impractical in high-volume, hour-long scenarios due to prohibitive computational costs.
For example, the Qwen2-VL-7B model requires approximately 40GB of GPU memory and 90 hours to handle a one-hour video frame-by-frame \cite{wang2024qwen2}. This makes these models unsuitable for high-volume video analytics.

In contrast, our method \textsc{Lava} employs strategic optimizations that enable it to process one-hour video streams in just a few minutes, making it exceptionally well-suited for efficient traffic video analytics even in high-volume scenarios.

\begin{figure*}
	\centering
	\begin{center}
		\includegraphics[width=\linewidth]{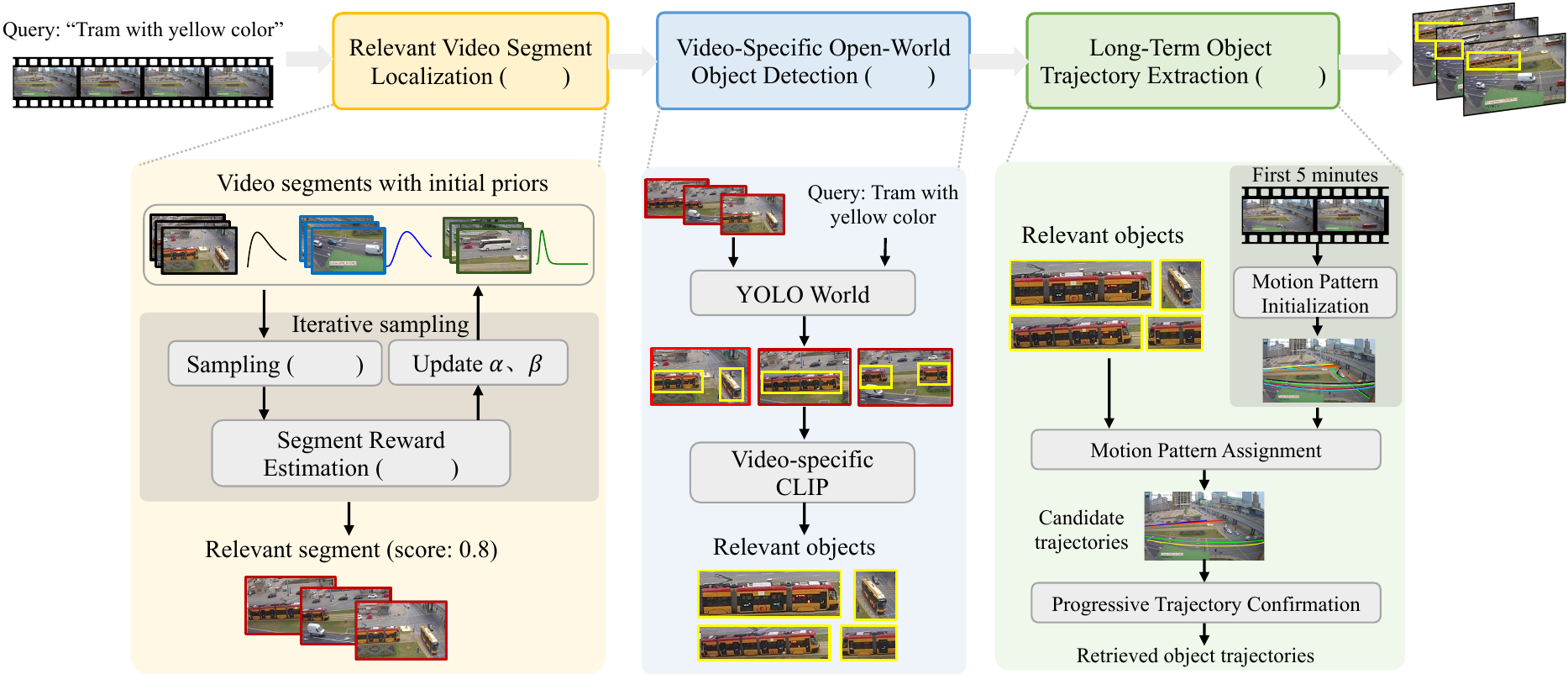}
		\put(-342, 189.5){\fontsize{9}{14}\selectfont{\S\ref{sec:Segment Localization}}}
		\put(-224, 189.5){\fontsize{9}{14}\selectfont{\S\ref{sec:Detection}}}
		\put(-98, 189.5){\fontsize{9}{14}\selectfont{\S\ref{sec:Motion}}}
		\put(-380.5, 64) {\fontsize{8}{14}\selectfont Eq. \ref{reward*}}
		\put(-410, 97) {\fontsize{8}{14}\selectfont Eq. \ref{eq_sampling}}
	\end{center}
	\vspace{-5pt}
	\caption{\small Illustration of \textsc{Lava} framework. \textsc{Lava} extracts object trajectories from high-volume videos specified in a language query  in a three-stage process:  (i) Relevant Video Segment Localization (\S\textcolor{red}{\ref{sec:Segment Localization}}) first equally partitions the video and initializes a prior for Thompson Sampling, which iteratively refines segment probabilities to identify those most likely to contain relevant objects; (ii) Video-Specific Open-World Object Detection (\S\textcolor{red}{\ref{sec:Detection}}) leverages YOLO World and a video-specific CLIP model to detect objects relevant to query; (iii) Long-Term Object Trajectory Extraction (\S\textcolor{red}{\ref{sec:Motion}}) initializes motion patterns from the first five minutes of video, then utilizes these motion patterns to link detections across frames, retrieving complete object trajectories of query targets. }
	\label{fig: framework}
	\vspace{-5pt}
\end{figure*}
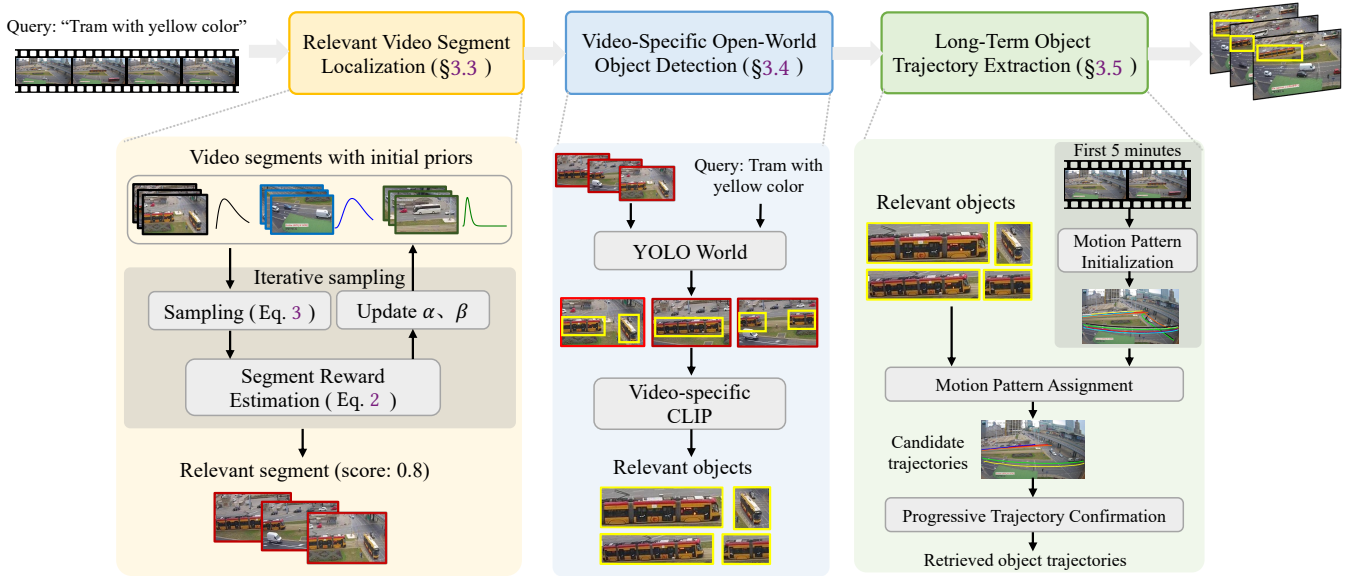

\section{Methodology}
\subsection{Task Description}
Language-driven scalable video analytics introduces a paradigm shift by allowing natural language predicates to define query tasks without being confined to predefined classes. This enables flexible, expressive, and precise video analytics.  Given a video $v$ and natural language query $p$, the task  needs to support three fundamental query types:

\begin{itemize}[leftmargin=*]
	\setlength{\itemsep}{0pt}
	\setlength{\parsep}{-2pt}
	\setlength{\parskip}{-0pt}
	\setlength{\leftmargin}{-10pt}
	\item \textbf{Selection queries:} These queries aim to retrieves all frames precisely matching the predicate $p$, such as, ``frames containing a black pickup truck with a white roof''.
	
	\item \textbf{Top-$k$ queries:} These queries seek to identify the top $k$ frames most relevant to the given description $p$. An example query is ``top-5 frames with multi-section buses'', where the system ranks frames based on semantic relevance to the description and returns the most pertinent ones.
	
	\item \textbf{Aggregation queries:} These queries compute  statistical summaries over the video. For instance, a query might ask for determining the average number of ``red sedans'' appearing across all frames. The system identifies relevant objects and calculates the required aggregation to produce the final result.
\end{itemize}

\subsection{System Architecture Overview}
As shown in Figure \ref{fig: framework}, \textsc{Lava} consists of three major components:  1) relevant video segment localization (\S\ref{sec:Segment Localization}) to quickly localize video segments relevant to queries; 2) video-specific open-world object detection (\S\ref{sec:Detection}) leverages YOLO-World and a video-specific CLIP to detect object-of-interests; 3) long-term object trajectory extraction (\S\ref{sec:Motion}) to determine complete video trajectories for the targets.

\subsection{Relevant Video Segment Localization}
\label{sec:Segment Localization}

\subsubsection{Multi-Armed Bandit Formulation}
We formalize relevant video segment localization as a multi-armed bandit problem \cite{osband2023approximate, galhotra2023metam, zhu2023scalable}. Particularly, we partition the video into equally sized segments and treat each segment as an independent ``arm''. Each segment’s reward is defined as the probability of containing query-relevant content. Our goal is to identify which segments are most relevant within a limited number of sampling rounds. We adopt Thompson Sampling~\cite{russo2018tutorial}, which balances exploration (sampling segments with greater uncertainty) and exploitation (prioritizing segments with higher expected rewards), enabling us to efficiently localize segments that align with the query predicate.

\subsubsection{Thompson Sampling}
To implement Thompson Sampling, we must first estimate the probability of each segment containing query-relevant content based on the existing observations. These estimates serve as guidance for Thompson Sampling, informing which segment to sample next and enabling a well-grounded balance between exploring less-certain segments and exploiting those more likely to contain the target objects.

\noindent\textbf{Segment Reward Estimation}
To determine which segment to sample next, given that $k$ samples have already been conducted, it is necessary to estimate $R_i(k+1)$ for each segment. Here, $R_i(k+1)$ represents the likelihood of identifying a frame containing objects relevant to the predicate $p$ in the next sample.

The estimation of $R_i(k+1)$ involves analyzing individual observations independently and aggregating insights for a global estimate. Suppose the video has $X$ frames containing the predicate $p$, divided into $L$ segments. Each segment $i$ contains $x_i$ frames with $p$, such that $\sum_{i=1}^{L}x_i = X$. For a segment $i$, the next sample's reward can be defined as:
{\small
\begin{equation}
	R_i(k+1) = \frac{x_i - x_{i,obs}}{n_i - n_{i,obs}},
\end{equation}}
where $x_{i,obs}$ denotes the cumulative frames in segment $i$ observed to contain $p$, $n_i$ is the total number of frames in $i$, and $n_{i,obs}$ is the number of sampled frames in $i$.

Since $x_i$ is unknown during the algorithm's execution, an iterative query-specific estimate is applied. The approximate reward, $R^*_i(k+1)$, is given by:
{\small
\begin{equation}\label{reward*}
	R^*_i(k+1) = \frac{x_{i,obs}}{n_{i,obs}}.
\end{equation}}
Unlike $x_i$, $x_{i,obs}$ is directly observable, enabling iterative refinement of estimates for effective segment selection.

\textbf{Theorem 1.} The approximate reward $R^*_i(k+1)$ is an unbiased estimate of $R_i(k+1)$, \ie., $E[R^*_i(k+1)]=E[R_i(k+1)]$. 

This result provides a solid basis for selecting chunks efficiently by reducing uncertainty and refining the sampling strategy using observed data. We compute $x_{i, obs}$ by applying the open-world object detector from Sec. \S\ref{sec:Detection} and counting the sampled frames where the query predicate is detected.

\noindent{\textbf{The Sampling Process}.}
 Building on the segment reward estimates, we now integrate these values into our sampling strategy to guide the choice of the next segment to sample. We model the potential reward distribution using a Gamma distribution, for each segment $i$, we configure the Gamma distribution with parameters $\alpha = x_{i, obs}$ and $\beta = n_{i, obs}$, ensuring its mean, $\alpha / \beta = x_{i, obs} / n_{i, obs}$, is consistent with the approximate reward $R_i^*(k+1)$ specified in Eq. (\ref{reward*}). This consistency forms the basis for using the Gamma distribution, enabling dynamic updates with new samples while preserving uncertainty in each segment’s estimated reward.

To handle cases where $\alpha$ or $\beta$ might be zero, we introduce small constants $\alpha_0$ and $\beta_0$ to ensure the stability of the Gamma distribution. The resulting distribution for $R_i^*(k+1)$ is therefore:

{\small
\begin{equation}\label{eq_sampling}
	R_i^*(k+1) \sim \Gamma(\alpha = x_{i, obs} + \alpha_0, \beta = n_{i, obs} + \beta_0).
\end{equation}}

By drawing samples from this distribution, a flexible selection process is enabled that leverages the uncertainty of each segment's reward rather than relying on a single point estimate. Once the maximum number of sampling rounds is reached, we employ the mean of the Gamma distribution, \ie., $x_{i, obs} / n_{i, obs}$, as the relevance score for each segment to guide the segment localization. 

\subsection{Video-Specific Open-World Object Detection}
\label{sec:Detection}
In open-world video scenarios, many queries refer to targets with fine-grained attributes. Many targets are often blurred or low-resolution. This poses great challenges for existing open-world detectors to produce accurate detections \cite{cheng2024yolo, 23Grounding}. Moreover, retraining these models for each unique query as \cite{kang2017noscope} is computationally prohibitive and therefore impractical for large-scale deployments. In this work, we propose a video-specific prompt tuning method which first mines video-specific training samples from existing detectors \cite{cheng2024yolo}, and then tunes CLIP  using these samples in an efficient manner to filter low-quality detection. 

\begin{figure}[!t]
	\includegraphics[width=\linewidth]{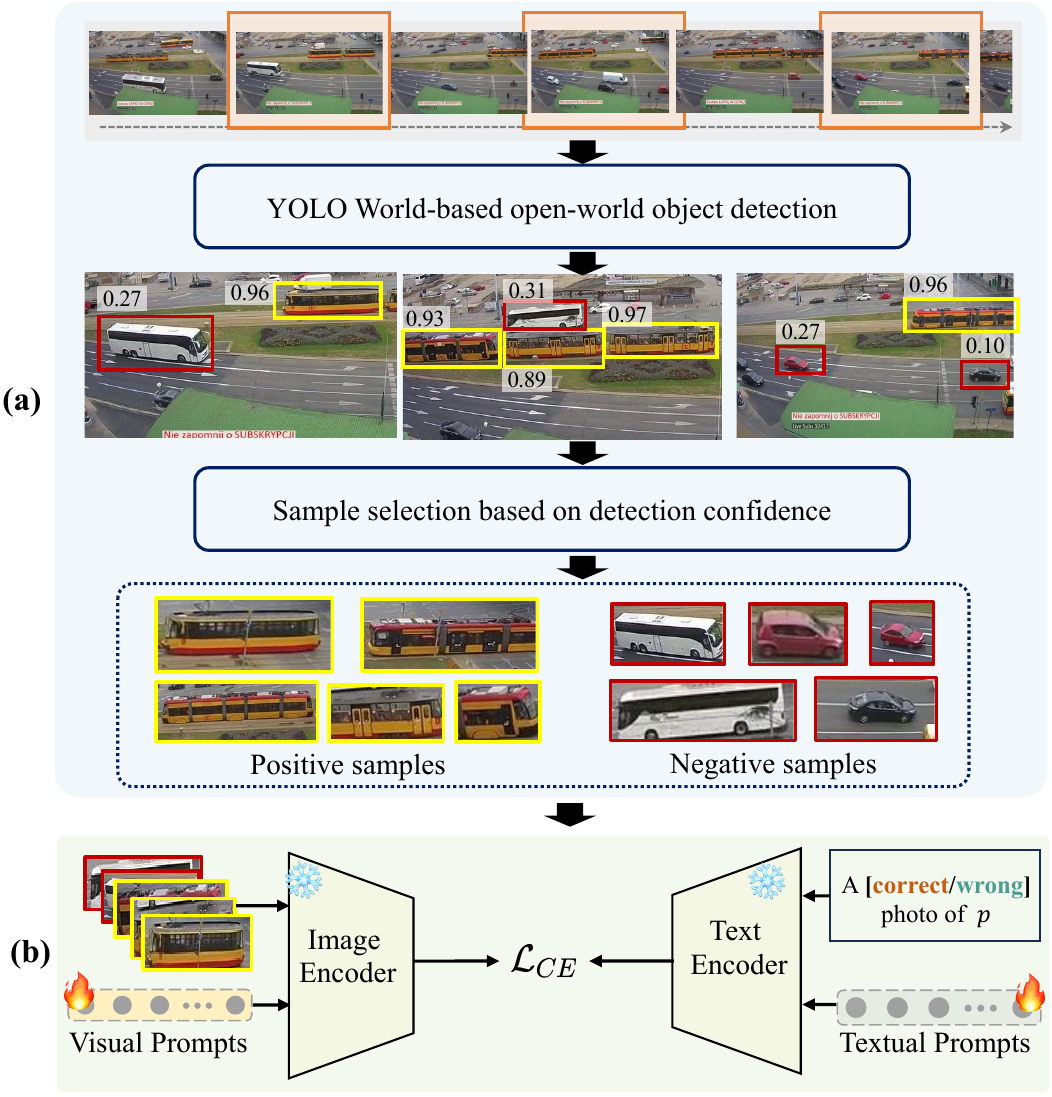}
	\caption{\small Illustration of (a) video-specific training sample mining and (b) prompt tuning. See \S\ref{sec:Detection} for details.}
	\label{fig: prompt}
	\vspace{-15pt}
\end{figure}

\subsubsection{Video-Specific Training Sample Mining}

As illustrated in Fig. \ref{fig: prompt} (a), for each video, we begin by uniformly sampling a small subset of frames across the entire video duration. This initial sampling step is designed to ensure a balanced temporal representation of the video content, capturing a diverse range of visual scenarios and object occurrences while substantially reducing the computational overhead associated with processing every frame. Subsequently, we perform open-vocabulary object detection on these sampled frames using YOLO-World \cite{cheng2024yolo}. This step involves obtaining  bounding boxes and associated confidence scores for each detected object, providing detailed spatial and semantic information. Next, we adopt a confidence-based selection strategy, wherein objects detected with sufficiently high confidence scores are categorized as positive samples, indicating a reliable match to the query predicate, while detections exhibiting low confidence are treated as negative samples, thus marking them as irrelevant or false detections relative to the specified predicate. This deliberate and selective filtering substantially enhances the reliability and quality of our training data by minimizing ambiguities and label noise, even though it reduces the overall dataset size. Notably, prior research has consistently demonstrated that prompt tuning methods achieve robust and effective performance even when trained with a relatively small number samples per class \cite{zhou2022coop, khattak2023maple, khattak2023self, chen2023retrospect, cho2023distribution,hou2025prompt}. Such data efficiency is particularly valuable in our open-world scenario, where acquiring extensive, fully annotated datasets for every conceivable query is impractical and prohibitively expensive, thereby enabling  video analytics in high volume scenarios.

\subsubsection{Prompt Tuning (PT)}
As illustrated in Fig. \ref{fig: prompt}(b), for video-specific prompt tuning, we introduce learnable prompt vectors in both visual and textual branches of VLMs. Our approach formulates a binary classification task for image regions identified by object detectors. We utilize simple templates (``a correct/wrong photo of $p$'') for the labels $c \in \{1, 0\}$, indicating if the region matches the query predicate $p$.

In the textual branch, we combine learned vectors ${\mathcal{T}_1, \dots, \mathcal{T}_h}$ with label-specific embeddings $\mathcal{T}(y_c)$ to form textual prompts $\tilde{\mathcal{T}}(y_c)$. Similarly, in the visual branch, we concatenate learned vectors ${\mathcal{I}_1, \dots, \mathcal{I}_n}$ with the image region $z$ embedding $\mathcal{I}(z)$ to form visual prompts $\tilde{\mathcal{I}}(z)$. We encode these prompts via the CLIP encoders to obtain visual features $\tilde{\mathbf{Z}}$ and textual features $\tilde{\mathbf{Y}}c$:
{\small
\begin{equation}
	\tilde{\mathbf{Z}} = \Phi{image}(\tilde{\mathcal{I}}(z)) \quad \text{and} \quad \tilde{\mathbf{Y}}c = \Phi{text}(\tilde{\mathcal{T}}(y_c)).
\end{equation}}
The classification probability is computed as:
{\small
\begin{equation}
	P(c | z) = \frac{\exp(\text{sim}(\tilde{\mathbf{Z}}, \tilde{\mathbf{Y}}c) / \gamma)}{\sum_{j \in \{0, 1\}} \exp(\text{sim}(\tilde{\mathbf{Z}}, \tilde{\mathbf{Y}}_j) / \gamma)},
\end{equation}}
where $\text{sim}(\cdot, \cdot)$ denotes the cosine similarity and $\gamma$ is the temperature parameter. The parameters are optimized using the standard cross-entropy loss: $\mathcal{L}_{\text{CE}}(c, z) = -\log P(c | z)$.

\subsection{Long-Term Object Trajectory Extraction}
\label{sec:Motion}
In open-world settings, factors like motion blur and occlusions can cause intermittent detection failures, making it difficult to maintain consistent trajectories over long spans. To address this issue, \textsc{Lava} introduces motion pattern assignment (derived from clustering object detections across frames \cite{zhang2023co, shao2023does}) to leverage temporal continuity, enhancing trajectory completeness.

\subsubsection{Motion Pattern Initialization}
\textsc{Lava} uses the first few minutes of video frames in an initial setup phase to extract object trajectories and establish motion patterns using an object tracking algorithm \cite{aharon2022bot}. Each trajectory $t_i \in T$ is represented as a temporal sequence of bounding box centers $\{(x_{k}^i, y_{k}^i)\}_{k=1}^{n_i}$, where $n_i$ denotes its length. 

To achieve more flexible and robust clustering results, we adopt the Fuzzy C-Means (FCM) algorithm \cite{bezdek1984fcm}. Unlike hard clustering methods used in \cite{xu2024predictive,bastani2022otif}, FCM introduces probabilistic memberships, enabling multi-cluster associations. This accommodates overlapping trajectories and similar motion patterns, resolving ambiguities in grouping complex trajectory data. After clustering, each medoid $\mu_j$ serves as the representative motion pattern for its cluster, encapsulating typical movement within the scene.

\subsubsection{Motion Pattern Assignment}
For a new detection $O_0^i$ at frame $f_0$ with center $(x_0^i, y_0^i)$, \textsc{Lava} selects $k$ candidate motion pattern  from initialized clusters. Each motion pattern $\mu_j$ contains sequential coordinates $\{(x^j_k, y^j_k)\}_{k=1}^{n_j}$ along its temporal span $n_j$. 

The assignment cost for assigning $O_0^i$ to a candidate trajectory $\mu_j$ is calculated as the minimum distance between the object’s position and the trajectory points. Let $(x_{a}^j, y_{a}^j)$ denote the $a$-th point on trajectory $\mu_j$. The assignment cost is computed as:
\begin{equation}\label{distance}
	d((x_0^i, y_0^i), \mu_j) = \min_{a \in [1, n_j]} \sqrt{(x_0^i - x^j_a)^2 + (y_0^i - y^j_a)^2}.
\end{equation}
The $k$ most promising trajectories, selected based on their assignment costs, are chosen as candidates. Among these, the trajectory with the smallest assignment cost is selected to guide the uniform sampling of $n$ frames within its temporal span for refinement.

\subsubsection{Progressive Trajectory Confirmation} After the initial detection $O_{0}^i$ and the sampling of $n$ frames, \textsc{Lava} employs a confirmation mechanism to refine the assigned trajectory. The sampled frames are processed to verify the presence of objects that match the query predicate $p$ and to ensure consistency with $O^i_0$.

For each sampled frame $f_m$ ($m = 1, \dots, n$), \textsc{Lava} utilizes YOLO-World for object detection and a prompt-tuned CLIP model to filter detections by confirming their match with predicate $p$. To ensure consistency with $O^i_0$, \textsc{Lava} applies a ReID model \cite{he2020fastreid} to compute similarity scores between each detection $D_i^m$ in $D^m$ and $O^i_0$. Among detections with similarity scores exceeding the threshold $\tau_{r}$, the detection with the highest similarity score is selected as the detection of object $i$ in frame $f_m$, denoted as $O^i_m$.

For each $O^i_m$, \textsc{Lava} assigns it to $k$ candidate motion pattern trajectories from the pre-initialized clusters based on the assignment cost defined in Eq. (\ref{distance}). Let $\mathcal{T}_m$ denote the set of $k$ candidate trajectories for detection $O^i_m$. After processing all $n$ sampled frames, the candidate trajectories from each frame are unified into a collective candidate set $\mathcal{T}_{\text{candidate}}$.

These candidate trajectories are then evaluated to identify the most consistent trajectory with the selected detections. Specifically, the trajectory with the smallest average assignment cost to all selected detections is chosen as the final assigned trajectory:
\begin{equation}
	\mu_{\text{final}} = \arg\min_{\mu_j \in \mathcal{T}_{\text{candidate}}} \frac{1}{|O_i|} \sum_{O_{m}^i \in O_i} d((x_{m}^i, y_{m}^i), \mu_j),
\end{equation}
where $O_i$ represents the set of all detections $O^i_m$ for object $i$, and $d((x_{m}^i, y_{m}^i), \mu_j)$ is the assignment cost between the detection's position and trajectory $\mu_j$.

To determine the temporal span of the trajectory for object $i$, \textsc{Lava} identifies the closest points on $\mu_{\text{final}}$ for all detections $O^i_m$ of object $i$. The temporal interval of the trajectory is calculated by extending the earliest and latest frames among these closest points. The trajectory for object $i$ is then represented as $\mathbb{T}_i = \langle f_i^s, f_i^e \rangle$, indicating the temporal extent of object $i$'s presence based on valid detections and the motion pattern of $\mu_{\text{final}}$. 

\begin{table}[t]
\caption{\small{Datasets and Query Predicates}}
\centering
\small
\setlength{\tabcolsep}{4pt}
\renewcommand{\arraystretch}{1.0}
\begin{tabular}{c|c|l|c}
\hline\thickhline
\bfseries Dataset & \bfseries FPS & \bfseries Query Predicate & \bfseries Selectivity \\
\hline\hline
\multirow{3}{*}{Caldot1} & \multirow{3}{*}{15}
  & Black Sedan & 0.52 \\
& & SUV with roof racks & 0.24 \\
& & Pickup truck with a flatbed & 0.16 \\
\hline
\multirow{3}{*}{Caldot2} & \multirow{3}{*}{15}
  & Semi-truck with a trailer & 0.17 \\
& & Black SUV & 0.16 \\
& & Black pickup truck & 0.04 \\
\hline
\multirow{3}{*}{Tokyo} & \multirow{3}{*}{10}
  & Bus with single-section design & 0.98 \\
& & White truck with black lettering & 0.60 \\
& & Multi-section articulated bus & 0.01 \\
\hline
\multirow{3}{*}{Amsterdam} & \multirow{3}{*}{30}
  & White boat with black roof & 0.96 \\
& & Boat with a long mast and sails & 0.42 \\
& & Black bicycle & 0.28 \\
\hline
\multirow{3}{*}{Jackson} & \multirow{3}{*}{30}
  & SUV without roof rack & 0.89 \\
& & SUV with roof rack & 0.37 \\
& & White semi-truck & 0.03 \\
\hline
\multirow{3}{*}{Warsaw} & \multirow{3}{*}{10}
  & Bus with multi-section design & 0.75 \\
& & Tram with yellow and red coloration & 0.52 \\
& & Bus with single-section design & 0.26 \\
\hline
\end{tabular}
\label{tab:dataset}
\vspace{-20pt}
\label{tab: dataset}
\end{table}

\section{Dataset Construction}
\subsection{Dataset} 
To robustly evaluate methods and establish a reproducible benchmark for future research, we construct a comprehensive evaluation framework comprising six video datasets extensively validated in prior video analytics studies \cite{xu2024predictive,cao2022figo,moll2022exsample,lai2021top,kang2022tasti,xu2022eva}. Specifically, we use the Tokyo and Warsaw datasets from Miris \cite{bastani2020miris}, the Amsterdam and Jackson datasets from BlazeIt \cite{kang13blazeit}, and the Caldot1 and Caldot2 from OTIF \cite{bastani2022otif}.
All these datasets consist of videos from fixed street-level cameras. Amsterdam captures activity at a riverside plaza. Caldot1 and Caldot2 capture highway activities with low resolution. Jackson Town captures downtown traffic during nighttime. In contrast, Tokyo and Warsaw capture city traffic junctions. Following the OTIF setting, each dataset has a one hour training subset for model training and a one hour test subset for evaluation. 

\subsection{Labeling Process}
\noindent\textbf{Step 1. Query Predicate Construction.}
To generate diverse natural-language query predicates in videos, we generate vehicle captions via GPT-4o \cite{hurst2024gpt}, filter semantic redundancies, then employ CLIP \cite{radford2021learning} as a classifier to assign captions to each annotated bounding box in the original dataset.  To ensure comprehensive selection, we adopt a systematic strategy prioritizing frequency coverage: descriptors are first ranked by occurrence frequency, which we will refer to as selectivity. We then perform stratified selection across high-, mid-, and low-selectivity intervals. 

\noindent\textbf{Step 2. Candidate Frame Selection.}
Considering a 30 fps one-hour video with 108,000 frames, manually annotating each frame is extremely time-consuming. Besides, objects in the video have strong temporal correlation. To reduce the annotation workload, we adopt an efficient strategy. We run the leading object-tracking method BoT-SORT \cite{aharon2022bot} on each video. Based on its results, for each vehicle, we select three frames as candidate frames.

\noindent\textbf{Step 3. Human Labeling.}
We recruit eight experienced annotators. To ensure the accuracy of the annotations, each frame was evaluated by three people, and only when at least two annotators agreed that a frame matched the query was the annotation considered valid. Using the 18 query predicates constructed in Step 1 (three predicates per dataset based on selectivity), we conducted the labeling process across six datasets, annotating an average of 5,000 frames per dataset. Each participant spent approximately 60 hours on labeling, totaling around 480 hours. Detailed information about these datasets and queries is provided in Table~\ref{tab: dataset}.

\begin{table*}
    \caption{\small{Selection Query Performance on multiple datasets. Our method (\textsc{Lava}) achieves an average $F_1$-score of 0.64, outperforming the second-best method by 0.14, while its runtime is only moderately higher than that of LEAP, whose accuracy is significantly lower. Best and second-best results are highlighted in bold and underlined, respectively.}}
    \vspace{-.5em}
    \centering
    \setlength{\tabcolsep}{3pt} 
    \begin{tabular}{l||cc|cc|cc|cc|cc|cc}
        \thickhline
        & \multicolumn{2}{c|}{\textsc{Lava}} 
        & \multicolumn{2}{c|}{LEAP} 
        & \multicolumn{2}{c|}{OTIF} 
        & \multicolumn{2}{c|}{YOLO-World} 
        & \multicolumn{2}{c|}{Grounding DINO} 
        & \multicolumn{2}{c}{CLIP} \\
        \clineB{2-13}{1.5} 
        \multirow{2}{*}[10pt]{Dataset} & $F_1$ & Time (s) & $F_1$ & Time (s) & $F_1$ & Time (s) & $F_1$ & Time (s) & $F_1$ & Time (s) & $F_1$ & Time (s) \\
        \hline\hline
        Caldot1    & \textbf{0.63} &130.5&0.26&32.5&0.23&246.0&0.34&1736.3&\underline{0.40}&14545.2&0.31&181.6 \\
        Caldot2    &\underline{0.38}&76.7&0.33&36.2&\textbf{0.40}&240.5&0.36&1726.1&0.26&14611.8&0.25&181.2 \\
        Tokyo      &\textbf{0.62}&118.5&0.25&46.4&\underline{0.54}&70.0&0.59&501.3&0.52&16464.9&0.43&119.2 \\
        Amsterdam  &\textbf{0.75}&250.6&0.13&85.0&\underline{0.55}&276.2&0.60&1896.8&0.57&25403.3&0.49&353.9 \\
        Jackson    &\textbf{0.72}&220.1&\underline{0.55}&31.2&0.39&310.9&0.44&2186.3&0.39&24969.6&0.30&369.0 \\
        Warsaw     &\textbf{0.75}&97.5&0.61&44.0&0.68&71.2&0.64&519.0&0.65&17147.9&\underline{0.72}&116.0 \\
        \hline
        Avg &\textbf{0.64} &149.0&0.36&45.9&0.46&202.5&\underline{0.50}&1427.6&0.46&18857.1&0.42&220.2 \\
        \hline
    \end{tabular}
    \label{tab: selection}
	\vspace{-10pt}
\end{table*}

\section{Experiment}
\subsection{Experimental Setup}
\noindent\textbf{Baselines.}
We compare \textsc{Lava} against five baselines covering diverse approaches: open-vocabulary detectors (YOLO-World \cite{cheng2024yolo}, Grounding DINO \cite{23Grounding}), efficient video analytics frameworks (OTIF \cite{bastani2022otif}, LEAP \cite{xu2024predictive}), and a foundational vision-language model (CLIP \cite{radford2021learning}). To enable language-driven analytics, we adapt OTIF and LEAP by replacing their original detectors with an open-vocabulary model. Methods like Zelda \cite{romero2023zelda} and SeeSaw \cite{moll2023seesaw} leverage CLIP's capabilities for tasks such as retrieving top-$k$ relevant frames or interactive image search. However, these approaches are designed for fundamentally different tasks compared to our work. While direct comparison is infeasible, we use CLIP as a baseline to represent CLIP-based matching in video analytics, highlighting its insights into language-driven selection despite lacking support for aggregation or top-k queries. 


\noindent\textbf{Metrics.}  We use standard metrics as  conventions \cite{kang2017noscope,kang13blazeit,bastani2020miris,bastani2022otif,xu2024predictive}:
\begin{itemize}[leftmargin=*]
	\setlength{\itemsep}{0pt}
	\setlength{\parsep}{-2pt}
	\setlength{\parskip}{-0pt}
	\setlength{\leftmargin}{-10pt}
	\item For selection queries, the ground truth result set comprises video frames that contain the query object and we use $F_1$-score as the evaluation metric.
	\item For aggregation queries, we adopt mean absolute percentage error (MAPE) to measure the estimation error.
	\item Regarding top-$k$ queries, we employ precision as the performance metric, which measures the fraction of correct results among the top  returned items.
\end{itemize}

\noindent\textbf{Implementation Details.}
In implementing \textsc{Lava}, YOLO-World \cite{cheng2024yolo} is used for prompt tuning, processing uniformly sampled training frames (one every 100) with a confidence threshold of 0.85 to generate high-quality pseudo-labels. The ReID model, FastReID \cite{he2020fastreid}, is trained once on trajectory data from the initialization phase across all datasets, creating a generic model for consistent object re-identification without fine-tuning. YOLO-World serves as the open-vocabulary detector for all queries in both training and inference. In addition, the Relevant Segment Localization stage employs Thompson Sampling with $I_{\text{max}} = 2000$ and divides the video into 500 segments. For the Trajectory Extraction stage, we set the number of sampled frames $n$ to 5 to balance accuracy and efficiency.
For baselines, we use the open-sourced code and largest pre-trained models of YOLO-World and Grounding DINO to ensure fine-grained target recognition. In OTIF and LEAP, we replace their object detectors with YOLO-World to enable language-driven video analytics. Since OTIF supports variable sampling rates, we set its rate to the closest runtime to \textsc{Lava} for fair comparison.


\subsection{Performance on Video Query Processing}
All experiments reported below are following the OTIF \cite{bastani2022otif} setting, where each dataset includes a one-hour training subset for model training and a separate one-hour test subset for evaluation.

\noindent\textbf{Selection Query.}  
The results in Table~\ref{tab: selection} show that \textsc{Lava} achieves an average $F_1$ of \textbf{0.64}, outperforming the next best by \textbf{0.14}. \textsc{Lava} ranks first on five of six datasets and second on Caldot2, demonstrating robust performance. Notably, \textsc{Lava} excels on challenging scenes (Amsterdam: \textbf{0.75}, Tokyo: \textbf{0.62}). In contrast, baselines such as OTIF (0.46) and YOLO-World (0.50) exhibit large fluctuations (\eg, LEAP drops to 0.13 on Amsterdam; CLIP to 0.25 on Caldot2), underscoring \textsc{Lava}'s stability under sparse and complex predicates. 

\begin{table}[ht]
\centering
\caption{\small{Top-$k$ Query Performance. Each dataset's result represents the average precision over all query predicates within that dataset. \textsc{Lava} achieves the highest precision, outperforming the second-best method by 6 percentage points on average, demonstrating its superior ability to retrieve the most relevant frames.}}
\setlength{\tabcolsep}{7pt}
\small
\renewcommand{\arraystretch}{1.0}
\begin{tabular}{l||ccccc}
\thickhline
Dataset & \textsc{Lava} & LEAP & OTIF & YOLO-W & G-Dino \\ \hline\hline
Caldot1   & \textbf{0.87} & 0.23 & 0.40 & 0.37 & 0.83 \\ 
Caldot2   & 0.83 & 0.80 & 0.93 & \textbf{0.97} & 0.80 \\ 
Tokyo     & 0.70 & \textbf{0.87} & 0.83 & 0.80 & 0.80 \\ 
Amsterdam & \textbf{0.97} & 0.13 & 0.90 & 0.87 & 0.93 \\ 
Jackson   & 0.87 & \textbf{0.90} & 0.70 & 0.87 & 0.37 \\ 
Warsaw    & \textbf{0.93} & 0.73 & \textbf{0.93} & 0.87 & 0.70 \\ \hline
Avg       & \textbf{0.86} & 0.61 & 0.78 & 0.79 & 0.74 \\ 
\thickhline
\end{tabular}
\label{tab:topk_results}
\end{table}

\begin{figure}
	\centering
	\includegraphics[width=\linewidth]{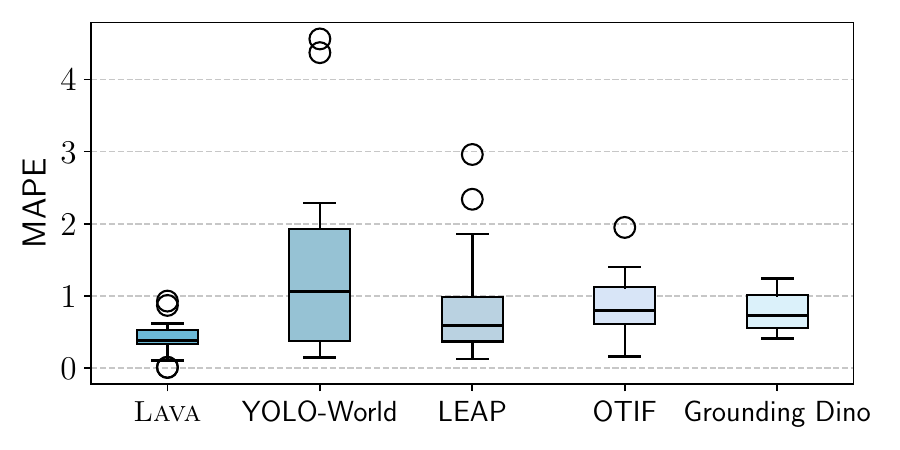}
	\caption{\small {Aggregation Query Performance over 18 Predicates. Box plots show the MAPE distribution: boxes mark the IQR with the median as the horizontal line, and circles are outliers. Lower and tighter distributions indicate better performance. \textsc{Lava} achieves the lowest median MAPE with minimal variability, demonstrating superior performance in aggregation queries.}}
	\label{fig: aggregation}
\end{figure}

\noindent\textbf{Aggregation Query.}
\textsc{Lava} reduces MAPE by $\mathbf{0.39}$ compared to the next best method, demonstrating superior aggregation query performance. As shown in Fig. \ref{fig: aggregation}, it achieves consistently lower MAPE with a narrow IQR, indicating high accuracy and stability. In contrast, YOLO-World has the highest variability, while LEAP, OTIF, and Grounding DINO show greater stability but higher MAPE, reflecting limited accuracy. These results highlight \textsc{Lava}'s effectiveness in aggregation queries, combining semantic filtering with reliable trajectory extraction to consistently achieve low MAPE values across diverse queries.

\noindent\textbf{Top-$k$ Query.}
As shown in Table~\ref{tab:topk_results}, \textsc{Lava} achieves the highest average precision (0.86), excelling in Caldot1, Amsterdam, and Warsaw. LEAP performs well in Tokyo and Jackson (0.87/0.90) but averages 0.61 overall. OTIF maintains 0.78 precision, particularly in Caldot2 and Amsterdam. YOLO-World peaks in Caldot2 (0.97) but averages 0.79, while Grounding-Dino excels in Amsterdam (0.93) but struggles in Jackson (0.74 average). These results highlight \textsc{Lava}'s robustness across datasets and complex queries where others falter.

\begin{figure}
	\centering
	\includegraphics[width=0.8\linewidth]{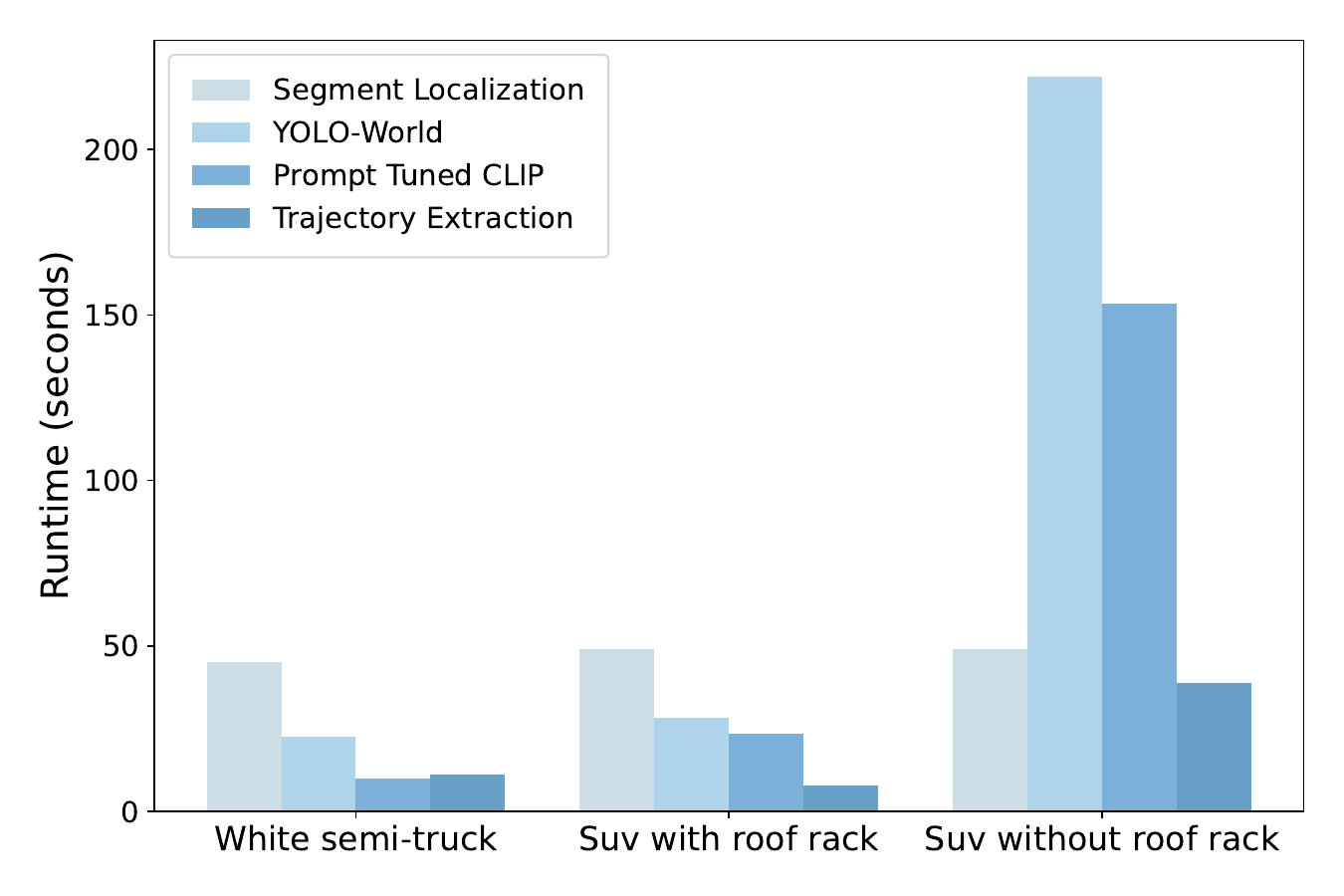}
	\caption{\small{Stage-wise runtime for three queries (White semi-truck, Suv with roof rack, Suv without roof rack) in the Jackson Town dataset with selectivities of 0.03, 0.37, and 0.89, representing low, medium, and high selectivity scenarios, respectively. The \textit{Segment Localization} stage shows minimal and fixed runtime, while the runtime of other stages increases as target selectivity becomes higher.}}
	\label{fig: execution_time}
\end{figure}

\noindent\textbf{Execution time Analysis.} 
In addition to its superior accuracy, \textsc{Lava} continues to demonstrate exceptional computational efficiency. The updated results show that \textsc{Lava} processes queries in an average time of $\mathbf{149.0\,\mathrm{s}}$, which is approximately \textbf{1/10} of the runtime required by the fastest OVD method, YOLO-World ($1427.6\,\mathrm{s}$), and vastly outperforms Grounding DINO ($18857.1\,\mathrm{s}$). Compared to OTIF and LEAP, which process queries in $202.5\,\mathrm{s}$ and $45.9\,\mathrm{s}$ respectively, \textsc{Lava}$'$s performance strikes a balance between speed and scalability, outperforming OTIF in efficiency while maintaining accuracy. 

igure~\ref{fig: execution_time} decomposes \textsc{Lava}'s runtime on Jackson Town into four stages: \textit{Segment Localization}, \textit{YOLO-World Detection}, \textit{Prompt-Tuned CLIP Filtering}, and \textit{Trajectory Extraction}. Owing to Thompson sampling, Segment Localization incurs minimal and invariant cost. The remaining stages scale with selectivity (0.03, 0.37, and 0.89), with YOLO-World Detection dominating overall runtime, Prompt-Tuned CLIP Filtering enhancing semantic precision, and Trajectory Extraction bridging temporal gaps via motion patterns. This analysis highlights \textsc{Lava}'s efficient allocation of computational resources under sparse query conditions.

\begin{table}[t]
	\centering
	\caption{\small{Ablation of \textsc{Lava}'s core components: Relevant Segment Localization (RSL), Video-Specific Detection (VSDet), and Long-Term Trajectory Extraction (LTT), with average performance across all predicates in each dataset.}}
	\setlength{\tabcolsep}{1pt}
	\begin{tabular}{@{}l|l||cccc@{}}
		\thickhline
		Dataset & Method & Selection & Top-$k$ & Aggregation & Time (s) \\ \hline\hline 
		\multirow{4}{*}{Caldot1} & \textsc{Lava} &\textbf{0.63}&\textbf{0.87}&\textbf{0.58}&\textbf{130.5}     \\ 
		& w/o RSL &0.61 &0.83 &0.59&332.7      \\
		& w/o VSDet&0.38&0.53&1.39&158.9\\
		& w/o LTT&0.53&0.77&0.61&236.5\\
		\hline
		\multirow{4}{*}{Caldot2} & \textsc{Lava} &\textbf{0.38}&\textbf{0.83}&\textbf{0.41}&\textbf{76.7}      \\ 
		& w/o RSL &0.35&0.80&0.49&488.6      \\ 
		& w/o VSDet&0.27&0.63&1.17&120.5\\
		& w/o LTT&0.32&0.77&0.63&327.6\\
		\hline
		\multirow{4}{*}{Tokyo} & \textsc{Lava} &\textbf{0.62}&0.70&\textbf{0.32}&\textbf{118.5}      \\ 
		& w/o RSL &0.62&\textbf{0.77}&0.53&268.2      \\ 
		& w/o VSDet&0.57&0.63&0.98&153.7\\
		& w/o LTT&0.59&0.67&0.61&197.5\\
		\hline
		\multirow{4}{*}{Amsterdam} & \textsc{Lava} &\textbf{0.75}&\textbf{0.97}&\textbf{0.42}&250.6      \\ 
		& w/o RSL & 0.73&\textbf{0.97}&0.76&572.5     \\
		& w/o VSDet&0.65&0.73&0.95&220.4\\
		& w/o LTT&0.70&0.87&0.72&465.3\\
		\hline
		\multirow{4}{*}{Jackson} & \textsc{Lava} &\textbf{0.72}&\textbf{0.87}&\textbf{0.57}&\textbf{220.1}      \\ 
		& w/o RSL &0.68&0.83&0.60&466.4      \\ 
		& w/o VSDet&0.54&0.73&0.84&257.4\\
		& w/o LTT&0.70&0.80&0.64&379.2\\
		\hline
		\multirow{4}{*}{Warsaw} & \textsc{Lava} &\textbf{0.75}&\textbf{0.93}&\textbf{0.18}&\textbf{97.5}      \\ 
		& w/o RSL &0.73&0.87&0.35&269.9      \\ 
		& w/o VSDet&0.62&0.83&0.86&115.7\\
		& w/o LTT&0.71&0.77&0.52&213.7\\
		\hline
	\end{tabular}
	\label{tab:ablation}
\end{table}

\subsection{Ablation Study}
\label{sec:ablation}

Table \ref{tab:ablation} ablates key components in \textsc{Lava}. Our key findings are: 

\noindent\textbf{Relevant Video Segment Localization (\S \ref{sec:Segment Localization})} reduces preprocessing time by 50-70\% through selective frame processing, \eg, decreasing from 332.7s to 130.5s on Caldot1, 488.6s to 76.7s on Caldot2, and 269.9s to 97.5s on Warsaw. This maintains precision while eliminating redundant computation.  It also ensures that downstream detection and trajectory extraction are focused only on semantically meaningful regions, enhancing overall system responsiveness.

\noindent\textbf{Video-Specific Open-World Object Detection (\S \ref{sec:Detection})} prevents accuracy degradation by removing irrelevant objects. Without this module, Selection and Top-$k$ scores drop significantly (Caldot1: 0.63 to 0.38 and 0.87 to 0.53; Warsaw: 0.75 to 0.62 and 0.93 to 0.83), while Aggregation MAPE increases by 140\% on Caldot1 (0.58 vs. 1.39) and 378\% on Warsaw (0.18 vs. 0.86). This highlights the importance of adapting the vision-language model to the video's unique visual context for robust semantic alignment.

\noindent\textbf{Long-Term Object Trajectory Extraction (\S \ref{sec:Motion})} enhances trajectory consistency and efficiency. Its absence degrades Selection (Amsterdam: 0.75 vs. 0.70; Jackson: 0.72 vs. 0.70) and increases MAPE (Amsterdam: 0.42 vs. 0.72; Jackson: 0.57 vs. 0.64). Preprocessing time also rises by 72\% on Jackson (220.1s vs. 379.2s) and 67\% on Tokyo (118.5s vs. 197.5s), demonstrating its dual role in accuracy and efficiency. By preserving temporal continuity, it extracts more complete object trajectories across long video spans.

Together, these results validate the complementary roles of these components in achieving efficiency and accuracy, enabling \textsc{Lava} to excel in diverse and complex video analytics.

\section{Conclusion}
In this work, we introduce \textsc{Lava}, a language-driven scalable video analytics system specifically designed for traffic surveillance. \textsc{Lava} addresses the critical limitations of existing methods, including restricted query flexibility and the lack of support for complex query types in traffic scenarios. By enabling natural language queries and seamlessly integrating advanced techniques like relevant segment localization, semantic filtering, and motion pattern assignment, \textsc{Lava} achieves state-of-the-art performance with significantly reduced computational overhead. Extensive experiments on our newly proposed benchmark demonstrate \textsc{Lava}'s strong performance across multiple tasks, firmly establishing it as a robust, efficient, and scalable solution for language-driven video analytics.



\bibliographystyle{ACM-Reference-Format}
\bibliography{lava}

\end{document}